\documentclass[lettersize,journal]{IEEEtran}
\usepackage{amsmath,amsfonts}
\usepackage{algorithmic}
\usepackage{algorithm}
\usepackage{array}
\usepackage[caption=false,font=normalsize,labelfont=sf,textfont=sf]{subfig}
\usepackage{textcomp}
\usepackage{stfloats}
\usepackage{url}
\usepackage{verbatim}
\usepackage{graphicx}
\usepackage{cite}
\usepackage{verbatim}
\usepackage{graphicx}
\usepackage{multirow}
\usepackage{amsthm}

\newcommand{\ie}{\textnormal {i.e.}}

\newtheorem{definition}{Definition}

\newtheorem{theorem}{\textbf{Theorem}}
\usepackage{adjustbox}
\usepackage{tikz}
\usepackage{booktabs}
\usepackage{enumitem}

\begin{document}

\title{MEASURE: Multi-scale Minimal Sufficient Representation Learning\\ for Domain Generalization in Sleep Staging}

\author{Sangmin Jo, Jee Seok Yoon, Wootaek Jeong, Kwanseok Oh, Heung-Il Suk,~\IEEEmembership{Senior Member,~IEEE,}

    \thanks{S. Jo, W. Jeong, K. Oh, and H.-I. Suk are with the Department of Artificial Intelligence, Korea University, Seoul 02841, Republic of Korea (e-mail: hisuk@korea.ac.kr).}%
    \thanks{J. S. Yoon is with the Department of Brain and Cognitive Engineering, Korea University, Seoul 02841, Republic of Korea. Correspondence: hisuk@korea.ac.kr (Heung-Il Suk) }
}

\maketitle

\begin{abstract}
Deep learning-based automatic sleep staging has significantly advanced in performance and plays a crucial role in the diagnosis of sleep disorders. However, those models often struggle to generalize on unseen subjects due to variability in physiological signals, resulting in degraded performance in out-of-distribution scenarios. To address this issue, domain generalization approaches have recently been studied to ensure generalized performance on unseen domains during training. Among those techniques, contrastive learning has proven its validity in learning domain-invariant features by aligning samples of the same class across different domains.  Despite its potential, many existing methods are insufficient to extract adequately domain-invariant representations, as they do not explicitly address domain characteristics embedded within the unshared information across samples. In this paper, we posit that mitigating such domain-relevant attributes—referred to as \textit{excess domain-relevant information}—is key to bridging the domain gap. However, the direct strategy to mitigate the domain-relevant attributes often overfits features at the high-level information, limiting their ability to leverage the diverse temporal and spectral information encoded in the multiple feature levels. To address these limitations, we propose a novel \textbf{MEASURE} (\textbf{M}ulti-scal\textbf{E} minim\textbf{A}l \textbf{S}Ufficient \textbf{R}epresentation l\textbf{E}arning) framework, which effectively reduces domain-relevant information while preserving essential temporal and spectral features for sleep stage classification. In our exhaustive experiments on publicly available sleep staging benchmark datasets, SleepEDF-20 and MASS, our proposed method consistently outperformed state-of-the-art methods. Our code is available at : https://github.com/ku-milab/Measure
\end{abstract}


\section{Introduction}
\IEEEPARstart{S}{leep} staging, the process of identifying and tracking transitions between different sleep stages over time, plays a pivotal role in analyzing sleep quality and treating sleep disorders \cite{scott2023emerging}. Typically, experts categorize sleep states into five stages—wake, N1, N2, N3, N4, and rapid eye movement (REM)— using polysomnography (PSG), which records various physiological signals such as electroencephalography (EEG), electrocardiography (ECG), and electromyography (EMG). While manual sleep staging remains the gold standard, it is both labor-intensive and time-consuming, often requiring trained specialists to examine hours of physiological data carefully. To alleviate these issues, deep learning (DL)-based techniques offer a powerful alternative by automating feature extraction and enabling accurate analysis of complex physiological signals. In particular, recent advancements in DL-based methods have achieved significant success by effectively leveraging EEG signals, which capture essential brain activity patterns for distinguishing between different sleep stages \cite{carskadon2005normal, lee2022dream, lee2024sleepyco}.

\begin{figure}[t!]
    \centering
    \includegraphics[width=0.47\textwidth]{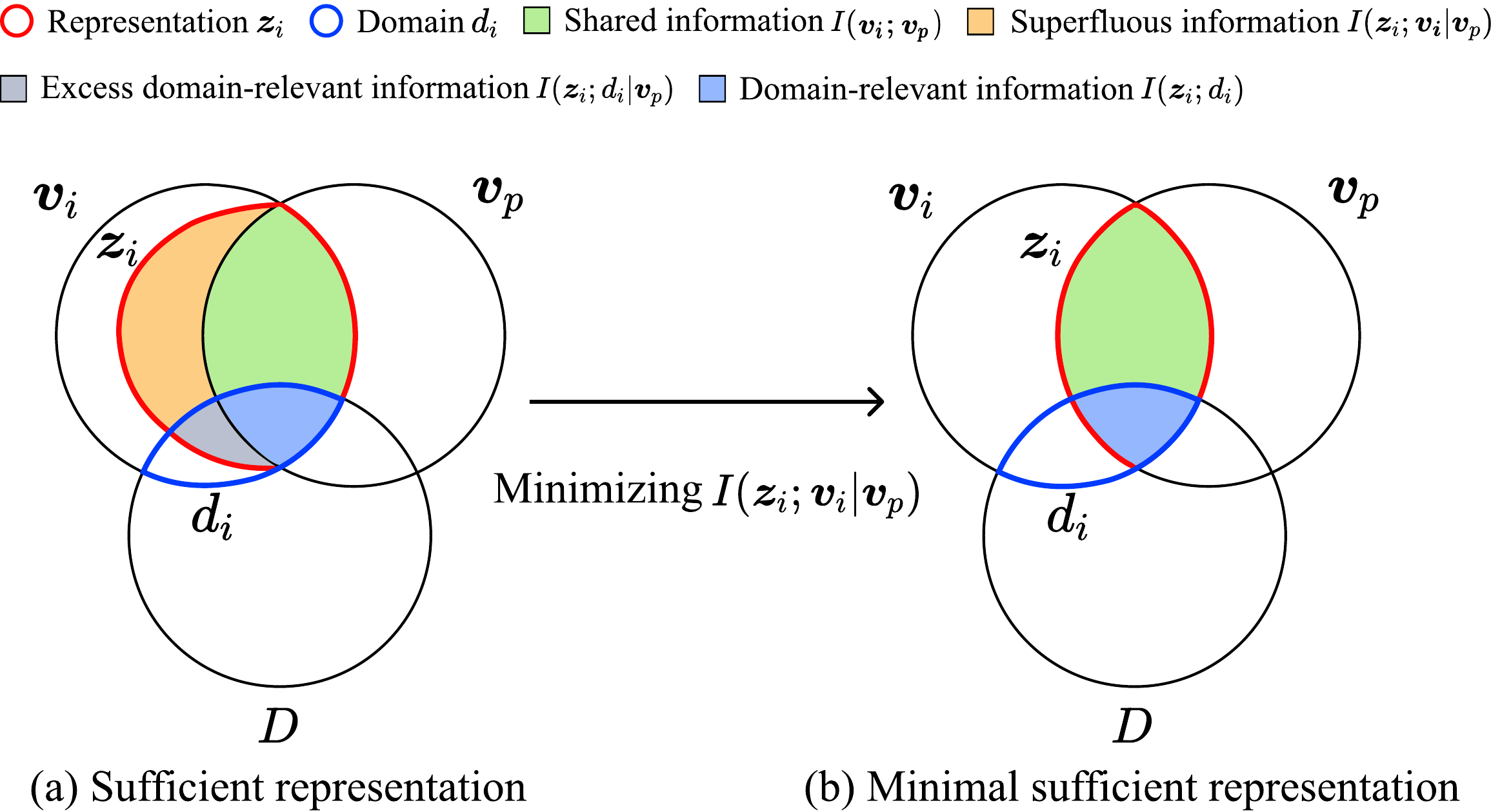} \\
    \caption{Comparison between (a) sufficient representation and (b) minimal sufficient representation. In conventional contrastive learning, $\boldsymbol{z}_i$ denotes the normalized feature of the $i$-th sample $\boldsymbol{v}_i$, while $\boldsymbol{z}_p$ represents the feature of a positive sample $\boldsymbol{v}_p$ that shares the same label as $\boldsymbol{v}_i$. The domain factor $D$ denotes the set of attributes that contribute to the domain gap. (a) Sufficient Representation Learning: This approach seeks to maximize the shared information between feature and positive samples $I(\boldsymbol{z}_i ; \boldsymbol{v}_p)$, while simultaneously introducing the superfluous information $I(\boldsymbol{z}_i; \boldsymbol{v}_i | \boldsymbol{v}_p)$, which corresponds to the information present in $\boldsymbol{v}_i$ but absent in $\boldsymbol{v}_p$. Among these, \textit{excess domain-relevant information} $I(\boldsymbol{z}_i ; d_i | \boldsymbol{v}_p)$ caused by domain attributes hinders the learning of domain-invariant features, where $d_i$ refers to the domain label of $\boldsymbol{v}_i$. (b) Minimal Sufficient Representation Learning: This approach aims to reduce the superfluous information $I(\boldsymbol{z}_i; \boldsymbol{v}_i | \boldsymbol{v}_p)$, thereby diminishing the excess domain-relevant information and enabling the learning of more domain-invariant features.
    }
    \label{fig:intro}
\end{figure}

Despite such advances, numerous DL-based techniques inevitably struggle when confronted with out-of-distribution (OOD) data (\ie, unseen subject or domain), leading to significant performance degradation caused by a discrepancy in data distribution \cite{zhou2022domain}. The challenge of OOD generalization in sleep staging is particularly prevalent due to the high variability in physiological signals among individuals. For instance, insomnia patients typically exhibit increased high-frequency activity and reduced slow-wave sleep in signals \cite{buysse2008eeg}. Moreover, age-related changes add to this complexity; research has shown that slow-wave sleep decreases with age—by as much as 2$\%$ per decade in adults—while the proportions of N2 and REM sleep undergo significant shifts across the lifespan \cite{ohayon2004meta}. These subject-specific characteristics pose a considerable challenge for DL models, often causing them to perform poorly on data from unseen subjects.

In this context, domain generalization (DG) aims to enhance the robustness of DL models by improving their ability to generalize across unseen data domains. Prior works in DG have focused on learning domain-invariant features by aligning multiple source domains~ \cite{li2018deep, mahajan2021domain,lu2022domain, dayal2024madg}. Within this paradigm, contrastive learning-based DG techniques have recently emerged as a promising strategy for extracting domain-invariant representation \cite{mahajan2021domain,yao2022pcl,liu2023promoting}. These methods effectively align multiple domains by clustering samples of the same label (\ie, positive pairs) from different domains while simultaneously pushing apart dissimilar ones (\ie, negative pairs). Owing to their advantageous properties, such approaches have demonstrated success in learning generalized representations from biosignals, suggesting their potential applicability in sleep staging \cite{zhang2022self, lee2022dream,jo2024enhancing,wang2024contrast}. They focus on increasing the information shared between positive samples, facilitating sufficient representation learning, where the learned features retain all task-relevant information \cite{tsai2021self}. However, such methods are likely to maintain superfluous information—unshared information across different samples~\cite{federici2020learning}—within the learned representations. Specifically, attributes arising from intra-class diversity, data augmentation artifacts, noise, and domain-specific traits may persist in the features.

In this work, we refer to the portion of superfluous information induced by domain gaps as \textit{excess domain-relevant information.} This information hinders the effective achievement of domain-invariant learning by embedding domain-specific characteristics within the feature space, as shown in Fig.~\ref{fig:intro}(a). In contrast, minimal sufficient representation learning reduces superfluous information during training, enabling the learning of more robust domain-invariant features, as illustrated in Fig.~\ref{fig:intro}(b). Hence, we leverage minimal sufficient representation learning to systematically reduce superfluous information, with a particular focus on minimizing \textit{excess domain-relevant information} by seamlessly decreasing the mutual information between features and domain-specific characteristics.

However, these approaches carry a potential risk of overfitting the features of the final encoder layer, as this may inadvertently decline the diversity of the learned representations. This phenomenon is particularly significant in sleep staging tasks because it is crucial to leverage multi-scale features from various layers of an encoder, which are capable of capturing diverse temporal and spectral scales, as highlighted in prior studies~\cite{wang2022novel, lee2024sleepyco}. To address these challenges, appropriately well-designed methods are required to eliminate excess domain-relevant information within multi-scale features.

To this end, we propose a novel framework called \textbf{M}ulti-scal\textbf{E} minim\textbf{A}l \textbf{S}Ufficient \textbf{R}epresentation l\textbf{E}arning (\textbf{MEASURE}), designed to leverage multi-scale domain-invariant features to effectively bridge distribution gaps. The primary objective of our MEASURE framework is to achieve robust domain generalization by minimizing domain discrepancies. Specifically, we extend minimal sufficient learning to a domain generalization setting, aiming to extract domain-invariant features by reducing excess domain-relevant information. We also provide a theoretical analysis of the proposed MEASURE framework, which not only highlights its ability to reduce domain discrepancies but also advances both the theoretical and practical understanding of domain generalization.

To further address the potential risks associated with reduced feature diversity in minimal sufficient learning, we enhance the proposed framework by extending the objective function to operate across encoder features extracted at multiple layers. This design ensures the model can effectively capture the diverse temporal and spectral characteristics inherent in sleep signals, thereby preserving information across feature hierarchies. Consequently, the main contributions of our work are:

\begin{itemize}
\item  To the best of our knowledge, we first introduce a theoretically grounded objective function for reducing excess domain-relevant information, offering a more effective approach for domain generalization compared to conventional contrastive learning methods.
\item We propose a novel integration of minimal sufficient representation learning within the multi-scale learning, effectively preventing overemphasis on specific layer features and enhancing generalization across domains.
\item We demonstrate the superiority of our MEASURE over state-of-the-art (SOTA) approaches on two sleep staging datasets, achieving significant improvements.
\end{itemize}

\section{Related Work}

In this section, we review previous work on domain generalization, sleep staging, and multi-view information bottleneck, and contextualize our contributions.

\subsection{Domain Generalization}
Domain generalization techniques have been introduced to enhance model performance on unseen domains~\cite{li2018domain, arjovsky2019invariant,xu2021fourier}. A common strategy is to learn domain-invariant representations by aligning samples from different source domains \cite{volpi2018generalizing,ding2022domain,liu2024cross}. For example,  \cite{yao2022pcl} utilized proxy-based contrastive learning to acquire domain-invariant representations by facilitating effective domain alignment. \cite{dayal2024madg} introduced margin-based adversarial learning that uses margin loss-based discrepancy to learn domain-invariant features. Building on these advancements, several studies have investigated the application of domain generalization to sleep staging tasks \cite{yang2023manydg, wang2024contrast}. 
For instance, \cite{yang2023manydg} proposed a novel framework that uses mutual reconstruction and orthogonal projection techniques to extract domain-invariant features, addressing subject variability.
\cite{wang2024contrast} proposed a hierarchical contrastive framework for medical time series, effectively capturing diverse information to achieve robust performance on unseen subjects.

While existing methods focus on domain-invariant features, they often overlook temporal and spectral information. In contrast, our MEASURE captures both while reducing domain-relevant information across multiple feature levels.

\subsection{Automatic Sleep Staging}
Conventional DL-based sleep staging approaches primarily focused on facilitating the effective modeling of both spatial and temporal patterns in the PSG \cite{tsinalis2016automatic,supratak2020tinysleepnet,phan2022sleeptransformer,phyo2022transsleep,ko2024eeg}. Recent studies have introduced techniques that enable models to learn representations across multiple scales of the encoder, effectively reflecting diverse temporal and spectral characteristics \cite{eldele2021attention, wang2022novel,lee2024sleepyco}. For example, \cite{eldele2021attention} developed a multi-resolution CNN leveraging varying filter widths to capture features across multiple scales effectively. \cite{wang2022novel} introduced a multi-scale dual attention network for exploring complex EEG-based sleep staging. Similarly, \cite{lee2024sleepyco} proposed SleePyCo, which employs contrastive learning and a transformer-based classifier that takes multi-level features as input. However, these methods often struggle to generalize effectively to unseen subjects or domains due to variability in physiological signals and environmental factors. To solve this problem, recent works have focused on domain generalization techniques \cite{jia2021multi,lee2022dream, wang2024generalizable}. For instance, \cite{jia2021multi} utilized adversarial learning with a domain classifier to improve generalization across diverse subjects. \cite{lee2022dream} employed a variational autoencoder and contrastive learning to disentangle domain-specific characteristics from features. \cite{wang2024generalizable} proposed a method for obtaining domain-invariant features through both epoch-level feature alignment and sequence-level alignment by treating datasets as domains.

While previous studies have sought to exploit multi-scale features or domain-invariant representations, they have struggled to effectively retain essential information within multi-scale features while ensuring robust domain invariance. In contrast, our study provides a theoretical rationale from the information bottleneck perspective and proposes a method to systematically mitigate domain-relevant information while preserving essential multi-scale representations.

\subsection{Multi-view information bottleneck}

In information bottleneck theory \cite{tishby2015deep, shwartz2017opening, tishby2000information}, robust representations are achieved by extracting task-relevant information while discarding irrelevant from the input. Based on this principle, multi-view information bottleneck (MVIB) studies seek to leverage the complementary nature of information across different augmented input from an input to improve representation learning \cite{federici2020learning,tsai2021self,wan2021multi,wang2022rethinking,wen2024mveb}. For example, \cite{federici2020learning} demonstrated that reducing superfluous information, which is information not shared across different views, is effective in enhancing representation learning. Similarly, \cite{tsai2021self} provides a theoretical grounding for multi-view-based self-supervised representation learning by discarding irrelevant features. \cite{wan2021multi} introduces a method to effectively integrate shared and view-specific information across multiple views using the information bottleneck principle in unsupervised multi-view representation learning. Building on these works, \cite{wen2024mveb} introduced an approach that extends mutual information to entropy and approximates it using a von Mises-Fisher distribution, demonstrating improved performance across benchmarks. 

Inspired by these studies, our method addresses the challenge of superfluous information from a domain-specific perspective, differentiating it from prior approaches that do not explicitly consider domain characteristics. Unlike the existing approaches, we introduce excess domain-relevant information as a novel type of domain-specific characteristics within superfluous information. By specifically focusing on this information, our approach enables the extraction of more robust domain-invariant representations, offering a novel perspective in solving DG challenges.

\section{Preliminaries}
Contrastive learning aims to learn robust representations by enhancing the similarity between multi-views of each sample. In this context, views refer to different augmentations applied to the same input sample. Let $\boldsymbol{v}_1$, $\boldsymbol{v}_2$, and $\boldsymbol{z}_1$, $\boldsymbol{z}_2$ represent two different views of the input sample $\boldsymbol{x}$ and normalized vectors of the projection head outputs from each view, respectively. Here, the projection head is typically a multi-layer perceptron to map low-dimension space. 

The contrastive loss ensures representation consistency by maximizing $I(\boldsymbol{z}_1 ; \boldsymbol{z}_2)$. Leveraging the data processing inequality~\cite{beaudry2012intuitive}, maximizing $I(\boldsymbol{z}_1 ; \boldsymbol{z}_2)$ serves as a lower bound for $I(\boldsymbol{z}_1 ; \boldsymbol{v}_2)$. Consequently, this framework enhances the mutual information $I(\boldsymbol{z}_1 ; \boldsymbol{v}_2)$ ensuring robust alignment between the learned representation and the augmented view~\cite{tsai2021self}.

\begin{figure*}[ht!]
    \centering
    \includegraphics[width=1\textwidth]{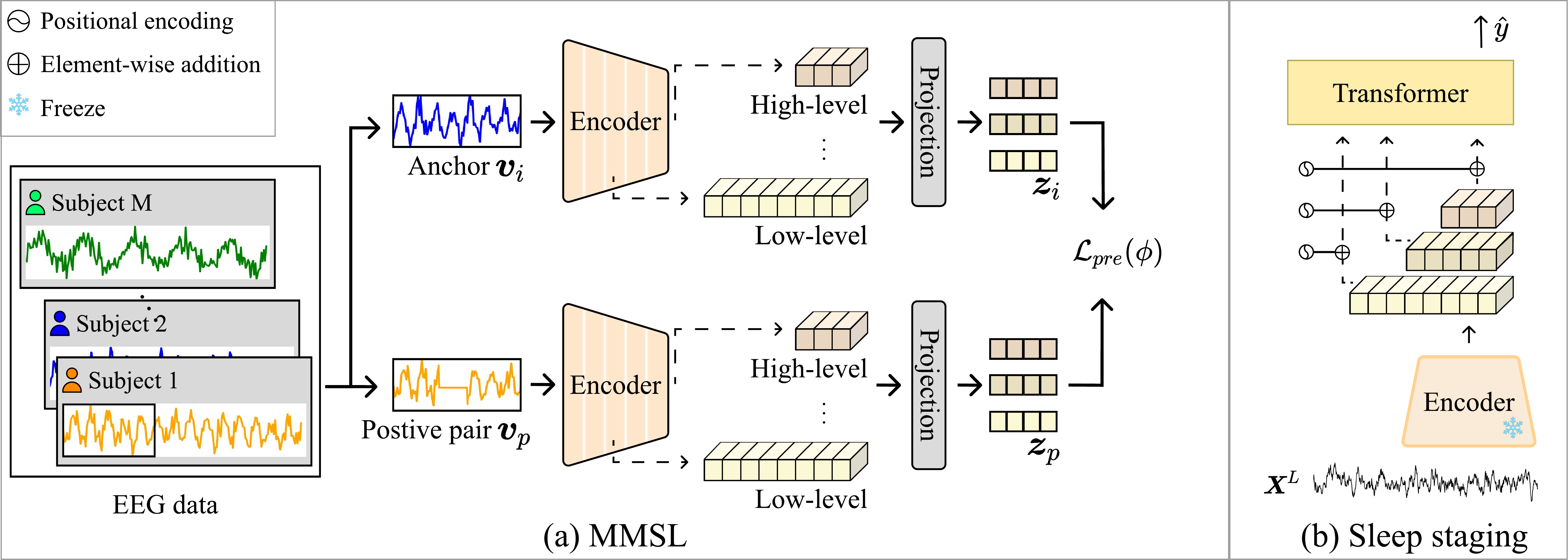}
    \caption{
    Overview of MEASURE. Our method consists of two stages: (a) Multi-scale minimal sufficient representation learning and (b) Sleep staging. In stage (a), multi-scale features are extracted from various layers of the encoder, capturing diverse frequency and temporal information. These features are subsequently projected into a shared feature space and optimized using the proposed objective. In the sleep staging (b), the encoder learned in the early stage is frozen, and the extracted multi-scale features are fed into a transformer to produce level-specific predictions. The final predicted sleep stage label $\hat{y}$ is obtained by aggregating these predictions using an argmax operation, as described in \cite{lee2024sleepyco}.
     }
    \label{fig:overall}
\end{figure*}

\begin{definition}[Sufficient representation for contrastive learning] \label{def: 1}
 A representation $\boldsymbol{z}_1^{suf}$ is considered sufficient for $\boldsymbol{v}_2$ if and only if $I(\boldsymbol{z}_1^{suf} ; \boldsymbol{v}_2) = I(\boldsymbol{v}_1 ; \boldsymbol{v}_2)$ 
\end{definition}

This definition implies that a sufficient representation $\boldsymbol{z}_1^{suf}$ preserves all the information that $\boldsymbol{v}_1$ contains about $\boldsymbol{v}_2$ ~\cite{wang2022rethinking}.  The sufficient representation $\boldsymbol{z}_1^{suf}$ inherently captures task-relevant features, as it is typically assumed that $\boldsymbol{v}_1$ and $\boldsymbol{v}_2$ share sufficient information for the task \cite{wen2024mveb}.

\begin{definition}[Minimal sufficient representation] A minimal sufficient representation $\boldsymbol{z}_1^{min}$ is considered minimal sufficient for $\boldsymbol{v}_2$ if and only if $I(\boldsymbol{z}_1^{min} ; \boldsymbol{v}_1 | \boldsymbol{v}_2) = 0$, for all sufficient representations.
\end{definition}

The \textit{superfluous information} refers to the information that is not shared between the two views, and it can be represented as conditional mutual information $I(\boldsymbol{z}_1 ; \boldsymbol{v}_1 | \boldsymbol{v}_2)$. A minimal sufficient representation $\boldsymbol{z}_1^{min}$ retains the least amount of this superfluous information for all sufficient representations. In previous studies, minimal sufficient representation learning has been demonstrated to enhance the robustness of representation learning \cite{wan2021multi, wen2024mveb}.

\section{Method}

\subsection{Problem Formulation} 
The goal of domain generalization in sleep staging is to train a model that generalizes to unseen target domains using only samples of source domains. It remains a challenging task due to the inherent domain shift problem, which is primarily attributed to inter-subject variability in EEG signals. 

To formally characterize this variability, we define the domain factor $D$ as the set of variables contributing to variability in EEG signals across different individuals, including but not limited to factors such as age, gender, and pathological conditions. Let $\mathcal{X}$ be the input space and $\mathcal{Y}$ be the label space. We denote multiple several domains as $\mathcal{D}_m := \{(\boldsymbol{x}_{k}^{m}, y_{k}^{m})\}_{{k}=1}^{N_m}$, where $ m \in \{1,2,3, \cdots, M\}$ denotes the $m$-th domain, $M$ is the number of domains, and $N_m$ is the number of samples in $m$-th domains. Here, $\boldsymbol{x}_{k}^{m} \in \mathbb{R}^{C\times T}$ represents the $k$-th EEG signal sample in $m$-th domain, where $C$ denotes the number of EEG channels, and $T$ represents the number of time points in the signal. The corresponding sleep stage label for the sample is denoted as $y_k^{m}$ while the domain label is represented as $d_k =m$ identifying the $m$-th domain within the domain factor $D$. We define $\boldsymbol{X}_{k}^{L}$ as a sequence that consists of the $k$-th sample along with the preceding $L$ samples, i.e., $ \{\boldsymbol{x}_{k-L},\boldsymbol{x}_{k-L+1} ,\cdots, \boldsymbol{x}_{k}\}$. 

The target domain $\mathcal{T}$ is defined as $\mathcal{D}_{m = U }$, and the source domain  $\mathcal{S}$ is defined as $\mathcal{D}_{m \neq U}$, where $U$ represents the index set corresponding to the unseen target subjects. The goal of domain generalization in the sleep staging task is to learn the mapping function $g : \mathcal{X} \rightarrow \mathcal{Y}$ that can accurately predict the sleep stage given a sequence of signals ($\boldsymbol{X}_{k}^{L}$) on unseen target domain $\mathcal{T}$, using only data from the source domains $\mathcal{S}$.

\subsection{Overview} 

We introduce a novel method MEASURE for domain generalization in 
sleep staging that addresses domain shift and leverages multi-scale features. The proposed method MEASURE consists of two stages, as illustrated in Fig. \ref{fig:overall}. 

During the pre-training phase, our approach follows the conventional contrastive learning paradigm to learn feature extractor $f(\cdot)$. Unlike prior approaches, MEASURE simultaneously aligns multi-scale features and maximizes the conditional entropy $H(\boldsymbol{z}|d)$. This prevents samples belonging to the same domain from clustering closely, thereby enabling the extraction of domain-invariant features. The derivation of the corresponding regularization term for conditional entropy maximization is detailed in Section \ref{section: minimal}. 

In the second stage, the encoder is frozen, and a sequence of $L$ biosignals is processed through the encoder to extract multi-scale features. These features are then passed into a transformer-based architecture to generate predictions. The model architecture and training strategy are based on prior work \cite{lee2024sleepyco}, which demonstrated strong performance by employing contrastive learning and multi-scale transformer architecture. Further details are provided in Section~\ref{section: sleepstaging}.

\subsection{Minimal Sufficient and Domain-Invariant Representation Learning} \label{section: minimal}

In contrastive learning-based DG, the feature space is typically encouraged to align samples of the same class across various domains by increasing their similarity. However, while those methods may provide a sufficient presentation, they do not necessarily ensure the learning of a minimal sufficient representation. As a result, excess domain-relevant information that is not shared between different domains often remains within superfluous information, thereby making it insufficient to achieve domain-invariant features. Therefore, we posit that reducing the excess domain-relevant information is crucial for enhancing the generalization capability of the learned representation. To this end, we employ minimal sufficient representation learning alongside the minimization of $I(\boldsymbol{z};d)$.

First, we formalize the relationship between minimal sufficient representation learning and domain invariance to provide a theoretical foundation for its validity.
\begin{theorem} \label{theorem: 1}
The minimal sufficient representation  $\boldsymbol{z}_1^{min}$ is more domain-invariant compared to the sufficient representation $\boldsymbol{z}_1^{suf}$ (proof in Appendix ~\ref{Appendix: a}).
\end{theorem}

\begin{equation} \label{eq: theorem1}
I(\boldsymbol{z}_1^{suf};d_1) \ge I(\boldsymbol{z}_1^{min};d_1)
 \end{equation}
Intuitively, this theorem holds because the superfluous information $I(\boldsymbol{z}_1 ; \boldsymbol{v}_1 | \boldsymbol{v}_2)$ encompasses domain-relevant information contained in $\boldsymbol{z}_1$. Thus, domain-invariant features can be effectively obtained through minimal sufficient representation learning.

In multi-view information bottleneck research, minimal sufficient representations are obtained by maximizing the alignment between different views, subject to the constraint of minimizing superfluous information using the Lagrangian multiplier method~\cite{federici2020learning}:
    \begin{equation} \label{eq: 1}
        \mathcal{L}(\phi) = \lambda_1 I(\boldsymbol{z}_1;\boldsymbol{v}_1|\boldsymbol{v}_2)-I(\boldsymbol{z}_1;\boldsymbol{v}_{2}),
    \end{equation}
where $\phi$ is the model parameter and $\lambda_1$ is a Lagrangian multiplier. Furthermore, we minimize the domain-relevant information $I(\boldsymbol{z}_1; d_1)$ to suppress the excess domain-relevant information within the superfluous information effectively, as described by the following objective:
    \begin{equation} \label{eq: unsupervised}
        \mathcal{L}(\phi) = \lambda_1 I(\boldsymbol{z}_1;\boldsymbol{v}_1|\boldsymbol{v}_2) + \lambda_2 I(\boldsymbol{z}_i;d)-I(\boldsymbol{z}_1;\boldsymbol{v}_{2}),
    \end{equation}
where $\lambda_2$ is another Lagrangian multiplier. This loss function can be viewed as an extension of minimal sufficient representation learning to the domain generalization paradigm. It minimizes superfluous information while focusing on excess domain-relevant information, thereby enabling the extraction of domain-invariant features.

This objective is extended to a supervised version to enable comparisons across diverse samples, similar to prior DG studies \cite{jeon2021feature,shen2022contrastive,yao2022pcl,jo2024enhancing}. We can extend Eq. (\ref{eq: unsupervised}) to a supervised setting as follows:
\begin{equation} \label{eq: 2}
        \mathcal{L}(\phi) = \lambda_1 I(\boldsymbol{z}_i;\boldsymbol{v}_i|\boldsymbol{v}_p) + \lambda_2 I(\boldsymbol{z}_i;d_i) -I(\boldsymbol{z}_i;\boldsymbol{v}_{p}),
\end{equation}
where $p$ denotes the indices of the positive pair for $i$-th sample in the batch. This objective encourages samples of the same class to cluster closely, making the feature space more discriminative while minimizing domain-specific information, thereby rendering the feature space domain-invariant.

However, computing mutual information is notoriously challenging due to the need to estimate high-dimensional probability distributions. Recent advances \cite{wen2024mveb} have addressed this challenge by approximating mutual information using the von Mises-Fisher (vMF) distribution, which is well-suited for modeling data constrained to a hypersphere. To leverage this approximation, we first decompose the mutual information in terms of entropy as follows (see Appendix \ref{Appendix:b.1}):

\begin{equation} \label{eq: lagrangian2}
    \mathcal{L}(\phi) = (\lambda_1+1)H(\boldsymbol{z}_{i}|\boldsymbol{v}_{p}) + (\lambda_2 -1)H(\boldsymbol{z}_{i}) -\lambda_2 H(\boldsymbol{z}_i|d_i).
\end{equation}

For computational efficiency and to ensure stability during the optimization process, Eq. (\ref{eq: lagrangian2}) can be simplified by setting the $\lambda_2 = 1$ and redefining $\lambda_1$ as $\lambda$, as follows:
\begin{equation} \label{eq: 3}
    \mathcal{L}(\phi) = (\lambda + 1) H(\boldsymbol{z}_{i}|\boldsymbol{v}_{p}) -H(\boldsymbol{z}_i|d_i).
\end{equation}
The validity of this simplification is empirically supported by experimental results, as illustrated in Fig. \ref{fig: ab2}.

Since the joint distribution $p(\boldsymbol{z}_i, \boldsymbol{v}_p)$ is unknown, directly calculating the conditional entropy $H(\boldsymbol{z_{i}} |  \boldsymbol{v_{p}})$ becomes intractable. Therefore, we employ a variational approximation $q_{\phi}(\boldsymbol{z}_i, \boldsymbol{v}_p)$ and derive the upper bound: 
\begin{align} 
H(\boldsymbol{z}_i|\boldsymbol{v}_p) &=-\mathbb{E}_{p(\boldsymbol{z}_i,\boldsymbol{v}_p)}[\log p(\boldsymbol{z}_i|\boldsymbol{v}_p)] \\
&\leq -\mathbb{E}_{p(\boldsymbol{z}_i,\boldsymbol{v}_p)}[\log q_\phi(\boldsymbol{z}_i|\boldsymbol{v}_p)] \label{eq: 6} .
\end{align}
Hence, minimization of Eq.~(\ref{eq: 3})  can be achieved through the following objective:
\begin{equation} \label{eq: 7}
    \bar{\mathcal{L}}(\phi) = -(\lambda + 1)\mathbb{E}_{p(\boldsymbol{z}_i,\boldsymbol{v}_p)}[\log q_\phi(\boldsymbol{z}_i|\boldsymbol{v}_p)] -H(\boldsymbol{z}_i|d_i).
\end{equation}

To approximate $\mathbb{E}_{p(\boldsymbol{z}_i,\boldsymbol{v}_p)}[\log q_\phi(\boldsymbol{z}_i|\boldsymbol{v}_p)]$, we adopt the vMF distribution as described in \cite{wen2024mveb}. The core concept is that normalized feature $\boldsymbol{z}$ resides on a hypersphere, and the conditional distribution $p(\boldsymbol{z}_i|\boldsymbol{v}_p)$ can be expressed in terms of the cosine similarity between $\boldsymbol{z}_i$ and $\boldsymbol{z}_p$. By using this approximation, we can optimize Eq.~(\ref{eq: 7}) by minimizing the following objective (see Appendix \ref{Appendix:b.2} for comprehensive details): 

\begin{equation} \label{eq: 8}
        \hat{\mathcal{L}}(\phi) = - \mathbb{E}_{p(\boldsymbol{z}_i, \boldsymbol{z}_p)}[\boldsymbol{z}_i^T \boldsymbol{z}_p] - \beta H(\boldsymbol{z}_i|d_i),
\end{equation}
where $\beta$ is the balance factor.

To compute the conditional entropy $H(\boldsymbol{z}_i|d_i)$ within the Eq. (\ref{eq: 8}), we adopted Stein gradient approximation \cite{li2017gradient}, as utilized in \cite{wen2024mveb}. Specifically, the gradient $\nabla_\phi H(\boldsymbol{z}|d)$ is approximated using the score function $\hat{\mathbf{G}}^{\mathrm{Stein}}_m$, and the model parameter is updataed by maximizing $\nabla_\phi H(\boldsymbol{z}|d)$. Further details are provided in Appendix~\ref{Appendix:entropy}. 

\subsection{Integration of Contrastive Learning and Minimal Sufficient Representation Learning}\label{sub:combine}

The aforementioned objective carries a potential risk of reducing the discriminative power of the features by inadvertently discarding class-relevant information within the superfluous information. To complement this, we incorporate a negative pair term that pushes samples from different classes farther apart. This approach encourages the feature space to become more distinguishable by increasing the separation between samples belonging to different classes. This integrated objective can be expressed as follows (more details in Appendix \ref{Appendix:b.3}):
\begin{align} \label{eq: cl}
    \tilde{\mathcal{L}}(\phi) &= \sum_{i \in I} \frac{-1}{|P(i)|} \sum_{p \in P(i)} \log \frac{\exp(\frac{\boldsymbol{z}_i^T \boldsymbol{z}_p}{\tau})}{\sum_{n \in N(i)} \exp (\frac{\boldsymbol{z}_i^T \boldsymbol{z}_n}{\tau})} \nonumber \\ 
    & \: \quad -\alpha  H(\boldsymbol{z}_i|d_i),
\end{align}
where $P(i) := \{p \in  A(i) \:| \:y_p = y_i \}$ denote the set of indices for positive pairs,  $N(i) :=  \{ n\in  A(i) \: | \: y_n \neq y_i \} $ is the set of indices of negative pairs for $i$-th instance in batch $A(i)$, $|P(i)|$ refer to cardinality of positive pair set, $\tau$ is the temperature parameter, and $\alpha$ is regularization parameter.

This objective function follows the form of conventional contrastive learning objectives while further enhancing domain-invariant properties by maximizing $H(\boldsymbol{z}|d)$. Moreover, the negative term exclusively considers samples from different classes, thereby making the feature space more discriminative.

\subsection{Preserving Multi-scale Features for Robust Sleep Staging} \label{section: multi-scale}
While minimal sufficient learning is crucial for mitigating domain gaps, this process may lead to overfit of features from specific layers due to reduced diversity of information. This phenomenon is particularly critical in sleep stage tasks, where multi-level features extracted from different encoder layers capture distinct frequency characteristics. For example, slow-wave sleep (N3) is associated with low frequencies (0.5–2 Hz), captured by lower-level features, while wake involves higher-frequency patterns (8–30 Hz), represented by higher-level features~\cite{berry2014aasm,lee2024sleepyco}. Therefore, it is essential to ensure that feature information across multiple levels is preserved while simultaneously extracting domain-invariant features.

To achieve this, we aim to employ minimal sufficient representation learning across multiple scales to effectively capture the diverse temporal and frequency characteristics present across different sleep stages. The objective for domain-invariant features in Eq. (\ref{eq: cl}) can be extended to account for multi-scale features as follows:
\begin{align} \label{eq: pre}
    {\mathcal{L}}_{\text{pre}}(\phi) &= \sum_{j\in J} \sum_{i \in I} \frac{-1}{|P(i)|} \sum_{p \in P(i)} 
    \log \frac{\exp(\frac{\boldsymbol{z}_{i,j}^T \boldsymbol{z}_{p,j}} {\tau})}{ 
\sum_{n \in N(i)} \exp (\frac{\boldsymbol{z}_{i,j}^T \boldsymbol{z}_{n,j}} {\tau})} \nonumber
    \\
    & - \alpha H(\boldsymbol{z}_{i,j}|d_i),
\end{align}
where $J$ represents the set of levels corresponding to the layers of the encoder, and $\boldsymbol{z}_{i,j}$ refers to the normalized feature from the output of the $j$-th layer for the $i$-th instance. This objective ensures that the model mitigates domain bias and avoids over-reliance on features from a specific layer. 

\subsection{Sleep Staging Phase} \label{section: sleepstaging}
In the sleep staging process, we employ SleePyCo \cite{lee2024sleepyco} as the backbone encoder, utilizing its transformer-based sequential classifier to predict sleep stages by leveraging multi-scale features. The backbone encoder, pre-trained using Eq. (\ref{eq: pre}), is frozen to preserve its learned domain-invariant and task-relevant features during this stage.

A $k$-th sequence composed of $L$ signal samples $\boldsymbol{X}_k^{L}$ is ed into the encoder to extract the features at the $j$-th level sequence features. These features are represented as $\boldsymbol{H}_{k,j}^{L} = \{ \boldsymbol{h}_{k-L, j}, \boldsymbol{h}_{k-L+1, j}, \dots, \boldsymbol{h}_{k, j} \}$, where $l$ denotes the length of the sequence corresponding to the $j$-th level feature extracted from the transformer. For each level $j$, these features are passed through a transformer to model temporal dependencies and obtain hidden states. The transformer's hidden states for $\boldsymbol{H}_{k,j}^{L}$ are aggregated using temporal attention, denoted as $\tilde{\boldsymbol{h}}_{k,j}$ to capture temporal dependencies effectively. Subsequently, these aggregated vectors are passed through linear layers to generate level-specific predictions $\boldsymbol{o}_{k,j}$. The final sleep stage prediction $\hat{y}_k$ is obtained by combining the outputs from all levels using $\hat{y}_k = \operatorname{argmax} \sum_j \boldsymbol{o}_{k,j}$. The detailed steps for the sleep staging process are outlined in Algorithm \ref{algorithm}.

\begin{algorithm}[!t] 
\footnotesize
\caption{Pseudo algorithm for the MEASURE}
\label{algorithm}
\begin{algorithmic}[1]
\REQUIRE{Training dataset $\mathcal{S}$, Augmentation module $\operatorname{Aug} (\cdot)$, Feature encoder network $f_{\phi}(\cdot)$, Projection head $\operatorname{Proj}_{\phi}(\cdot)$, Transformer $\operatorname{Tr}_{\psi}(\cdot)$, Temporal Attention module $\operatorname{TA}_{\psi}(\cdot)$, Linear layer $\operatorname{FC}_{\psi}(\cdot)$, Learning parameters $\phi_*$ and $\psi_*$, Cross-entropy loss $\text{CE}(\cdot)$, Stein gradient estimator $\operatorname{SGE}(\cdot)$, Regularization parameter $\alpha$, Learning rate $\eta$, Sequence length $L$, Multi-scale feature level $J$.
}

\textbf{Pre-training phase}\\
\FOR{$(\boldsymbol{x}, y, d)$ sampled from $S$ until convergence}
        \STATE{$\boldsymbol{v} = \operatorname{Aug}(\boldsymbol{x})$ }
        \STATE{$\boldsymbol{r}= f_{\phi}(\boldsymbol{v})$} \quad // $\boldsymbol{r}$ is multi-scaled features
        \FOR{each scale $j \in J $}
            \STATE{$\boldsymbol{z}_j = \operatorname{Proj}(\boldsymbol{r}_j)$} 
    
        \FOR{each domain $m = 1, \dots, M$}
            \STATE{$\hat{G}_{j,m}^{\text{Stein}} = \mathbb{E}_{p(\boldsymbol{z}_j|d=m)} [ \text{SGE}(\boldsymbol{z}_j) ]$  \quad // Compute Stein gradient}
            \STATE{ $H_j = -\mathbb{E}_{p(d)}[\hat{G}_{j,m}^{\text{Stein}} \cdot \boldsymbol{z}_j]$ \quad // Compute conditional entropy}
    \ENDFOR
    \ENDFOR
        \STATE{Calculate the pre-training loss using Eq.~(\ref{eq: pre})}
        \STATE{Update the encoder parameter $\phi$: }
        \STATE{$\phi \leftarrow \eta\nabla_{\phi} \mathcal{L}_\text{pre}$}
\ENDFOR
\RETURN{Trained multi-scale encoder network $f_\phi(\cdot)$}\\

\textbf{Sleep staging phase}\\
    \FOR{$\boldsymbol{X}^{L}$, $y$  sampled from $S$ until convergence}
    \STATE{$\hat{\boldsymbol{r}}^{L} = f_{\phi}(\boldsymbol{X}^{L})$}
        \FOR{each scale $j \in J $}
        \STATE{$H_j^{L} = \operatorname{Tr}_{\psi}(\boldsymbol{\hat{r}}^{L}_j)$}
        \STATE{$\hat{\boldsymbol{h}}_j = \operatorname{TA}_{\psi}(H_j^{L})$}  // reduce time dimension  \\
        \STATE{$\boldsymbol{o}_j = \operatorname{FC}_{\psi}(\hat{\boldsymbol{h}}_j )$}
        \ENDFOR
        
        \STATE{$\hat{y} = \operatorname{softmax} \sum_j \boldsymbol{o}_j$ }

        \STATE{$\mathcal{L}_{ce} = \text{CE}(y,\hat{y})$} 
        \STATE{Update the encoder parameter $\psi$: }
        \STATE{$\psi \leftarrow \eta\nabla_{\psi} \mathcal{L}_\text{ce}$}
        \ENDFOR

\RETURN{Trained transformer based classifier $\operatorname{Tr}_\psi(\cdot)$, $\operatorname{TA}_\psi(\cdot)$, and $\operatorname{FC}_{\psi}(\cdot)$}
\end{algorithmic}
\end{algorithm}

\section{Experiment}
\subsection{Dataset}

\begin{table*}[t!]

\caption{Performance comparison between our method and sleep staging SOTA methods, and DG approaches for sleep staging on SleepEDF-20 and MASS datasets. We evaluated performance using three metrics: Cohen’s Kappa ($\kappa$), accuracy (ACC), and macro-averaged F1 score (F1). Here, bold and underline indicate the best and second-best results, respectively. Values marked with  $^\dagger$ are reported from the original papers.}
\centering
\label{table:comparison}

\begin{tabular}{c c c ccc ccccc}
\toprule
\multirow{2}{*}{Datasets} & \multirow{2}{*}{Backbone Model} & \multirow{2}{*}{Method}  & \multicolumn{3}{c}{Overall metrics} & \multicolumn{5}{c}{Per-class F1 ($\%$)} \\
 \cmidrule(lr){4-6} 
& & & \multicolumn{1}{c}{$\kappa$} & \multicolumn{1}{c}{ACC ($\%$)} & \multicolumn{1}{c}{F1 ($\%$)}  & Wake & N1 & N2 & N3 & REM \\
\midrule
\midrule

\multirow{9}{*}{SleepEDF-20} & \multirow{4}{*}{Non-SleePyCo}

& $\text{IITNet}^\dagger$ \cite{seo2020intra} & 0.780 & 83.9 & 77.6 & 87.7 & 43.4 & 87.7 & 86.7 & 82.5 \\
& & SleepDG \cite{wang2024generalizable}   & 0.792 & 84.8 & 78.4 & 89.4 & 43.2 & 87.4 & \textbf{89.1} & 82.7   \\
& & $\text{Regularized SeqSleepNet}^\dagger$ \cite{phan2023improving}  & 0.811 & 86.2 & 79.3 & \underline{91.8} & 45.7 & 88.3 & 86.9 & 84.0 \\
& & $\text{XSleepNet}^\dagger$ \cite{phan2021xsleepnet}  & \underline{0.813} & \underline{86.3} & \underline{80.6} & - & - & - & - & - \\

 \cmidrule(lr){2-3}\cmidrule(lr){4-11}
 & \multirow{5}{*}{SleePyCo}
& IRM \cite{arjovsky2019invariant}  & 0.783 & 84.2 & 77.4 & 89.0 & 42.0 & 87.1 & 85.6 & 83.4 \\
& & PCL \cite{yao2022pcl}  & 0.809 & 86.0 & 80.1  & 90.1 & 48.3 & \underline{88.7} & 87.5 & 85.8 \\
& & SleePyCo (Base) \cite{lee2024sleepyco} & 0.812 & 86.2 & \underline{80.6} & 90.7 & \underline{50.0} & \underline{88.7} & 87.1 & \underline{86.3} \\
& & MEASURE (Ours) & \textbf{0.826} & \textbf{87.3} & \textbf{81.5}  & \textbf{92.6} & \textbf{50.4} & \textbf{89.3} & \underline{88.8} & \textbf{86.4} \\

\cmidrule(lr){1-11}
\multirow{9}{*}{MASS} & \multirow{4}{*}{Non-SleePyCo}

&  $\text{IITNet}^\dagger$ \cite{seo2020intra}  & 0.794 & 86.3 & 80.5 & 85.4 & 54.1 & 91.3 & 86.8 & 84.8 \\
& & SleepDG \cite{wang2024generalizable}  & 0.778 & 85.1 & 77.9 & 85.1 & 43.3 &90.1 & 87.7 & 82.6 \\
& & $\text{SleepMG}^\dagger$ \cite{ma2024sleepmg}& 0.802 & 86.6 & 81.7 & 85.1 & 43.3 & 90.9 & 87.7 & 82.6 \\
& & $\text{ProductGraph}^\dagger$ \cite{einizade2023productgraphsleepnet}  & 0.802 & 86.7 & 81.8  & \textbf{89.4} & 58.3 & 90.4 & 81.3 & \textbf{89.8} \\
 \cmidrule(lr){2-3}\cmidrule(lr){4-11}
& \multirow{5}{*}{SleePyCo} 
&  IRM \cite{arjovsky2019invariant}& 0.817 & 87.7 & 82.5 & 87.4 & 57.9 & 92.5 & \textbf{88.7} & 86.1 \\
& & PCL \cite{yao2022pcl}& 0.819 & 87.9 & \underline{82.9} & 88.0 & \underline{60.1} & 92.4 & 87.7 & 86.5 \\
& & SleePyCo (Base) \cite{lee2024sleepyco} & \underline{0.821} & \underline{88.0} & 82.8 & 86.5 & 59.4 & \textbf{92.8} & 88.1 & 87.5 \\
& & MEASURE (Ours) & \textbf{0.826} & \textbf{88.3} & \textbf{83.6} & \underline{88.2} & \textbf{61.3} & \underline{92.6} & \underline{88.2} & \underline{87.6} \\

\bottomrule
\end{tabular}
\end{table*}

We evaluated the performance of our proposed method on two different sleep staging datasets: SleepEDF-20 \cite{kemp2000analysis} and Montreal Archive of Sleep Studies (MASS) \cite{o2014montreal}.
The SleepEDF-20 dataset comprises PSG recordings from 20 subjects aged from 25 to 34. MASS contains PSG recordings from 62 subjects aged from 25 to 69. For the SleepEDF-20 dataset, we extracted a single-channel EEG (Fpz-Cz) sampled at 100Hz. We cropped the sleep recordings to ensure a 30-minute wake period before and after each recording. For the MASS dataset, we utilized the F4-LER channel, downsampled to 100Hz. For both datasets, we combined the N3 and N4 stages into a single N3 stage. This process is a commonly used data preprocessing method in sleep staging, and we adhered to the settings of numerous previous studies to ensure a fair comparison~\cite{seo2020intra, phan2021xsleepnet, phyo2022transsleep, lee2024sleepyco}. The class distribution of two datasets is in Table \ref{table: dataset}.

\subsection{Implementations Details} 
The model was pre-trained with a batch size of 1024, an initial learning rate of $3 \times 10^{-4}$, and a weight decay of $1 \times 10^{-4}$ for the Adam optimizer. To ensure a sufficient number of samples per domain for accurate computation of conditional entropy $H(\boldsymbol{z}|d)$ using the Stein gradient approximation, each batch was randomly constrained to contain samples from only two domains. The temperature hyperparameter $\tau$ for the contrastive loss was set to 0.07, while the regularization parameter $\alpha$ was set to 0.001. The sleep staging process follows the same architecture as the transformer-based classifier utilizing multi-scale features, as proposed in SleePyCo. For sleep staging, the pre-trained encoder was frozen, and only the classifier was trained, with the sequence length set to $L = 10$.

We employed the widely adopted k-fold cross-validation protocol to evaluate the performance of domain generalization. For each fold, we designated specific unseen subjects as the test set and repeated the experiment, ensuring that each subject was included in the test set exactly once. For the SleepEDF-20 dataset (k = 20), we partitioned the data into training, validation, and test sets with a ratio of 15:4:1, respectively. For the MASS dataset (k = 31), we used a ratio of 45:15:2 for training, validation, and test sets. All experiments were conducted on a server equipped with an NVIDIA RTX A6000 D6 48GB GPU.

\begin{table}[t]
\caption{Sleep stage distribution for SleepEDF-20 and MASS datasets.} 
\label{table: dataset}
\centering
\begin{tabular}{c cc} 
\toprule
Sleep stage & SleepEDF-20 & MASS \\
\midrule
\midrule
W & 8285 (19.6 \%) & 6231 (10.6 \%) \\
N1 & 2804 (6.6 \%) & 4814 (8.2 \%) \\
N2 & 17799 (42.1 \%) & 29777 (50.4 \%) \\
N3 & 5703 (13.5 \%) & 7653 (12.9 \%) \\
REM & 7717 (18.2 \%) & 10581 (17.9 \%) \\
Total & 42308 & 59056 \\
\bottomrule
\end{tabular}
\end{table}

\begin{table*}[h]
\caption{Comparison of MEASURE with DG methods and a contrastive learning-based baseline in DG evaluation. The symbol $\pm$ denotes the standard deviation.}
\label{table:DG}
    \centering
    \begin{tabular}{lccc|ccc}
        \toprule
        \multirow{2}{*}{Method} & \multicolumn{3}{c|}{SleepEDF20} & \multicolumn{3}{c}{MASS} \\
        \cmidrule(lr){2-4} \cmidrule(lr){5-7}
        & $\kappa$ & Acc (\%) & F1 (\%) & $\kappa$ & Acc (\%) & F1 (\%) \\
        \midrule
        \midrule
        IRM \cite{arjovsky2019invariant} & 0.787 $\pm$ 0.02 & 84.3 $\pm$ 1.54 & 77.4 $\pm$ 1.38 & 0.771 $\pm$ 0.03 & 84.7 $\pm$ 2.01 & 77.8 $\pm$ 3.17 \\
        SleepDG \cite{wang2024generalizable} & 0.798 $\pm$ 0.02 & 85.2 $\pm$ 1.26 & 78.0 $\pm$ 1.04 & 0.778 $\pm$ 0.01 & 85.1 $\pm$ 0.55 & 78.1 $\pm$ 1.10 \\
        PCL \cite{yao2022pcl} \cite{yao2022pcl} & \underline{0.811 $\pm$ 0.02} & \underline{86.1 $\pm$ 1.31} & 79.2 $\pm$ 1.05 & \underline{0.807 $\pm$ 0.03} & \underline{87.1 $\pm$ 1.82} &  \underline{82.0 $\pm$ 2.60} \\
        SleePyCo (Base) \cite{lee2024sleepyco} & 0.809 $\pm$ 0.02 & 85.8 $\pm$ 1.57 & \underline{79.6 $\pm$ 1.52} & \underline{0.807 $\pm$ 0.03} & \underline{87.1 $\pm$ 1.91} & 81.8 $\pm$ 2.75 \\
        Ours & \textbf{0.824 $\pm$ 0.01} & \textbf{87.0 $\pm$ 0.98} & \textbf{80.6 $\pm$ 1.05} & \textbf{0.810 $\pm$ 0.03} & \textbf{87.5 $\pm$ 1.67} & \textbf{82.7 $\pm$ 2.47} \\
        \bottomrule
    \end{tabular}
\end{table*}

\subsection{Results}
\begin{figure}[h] 
    \centering 
    \includegraphics[width=0.45\textwidth]{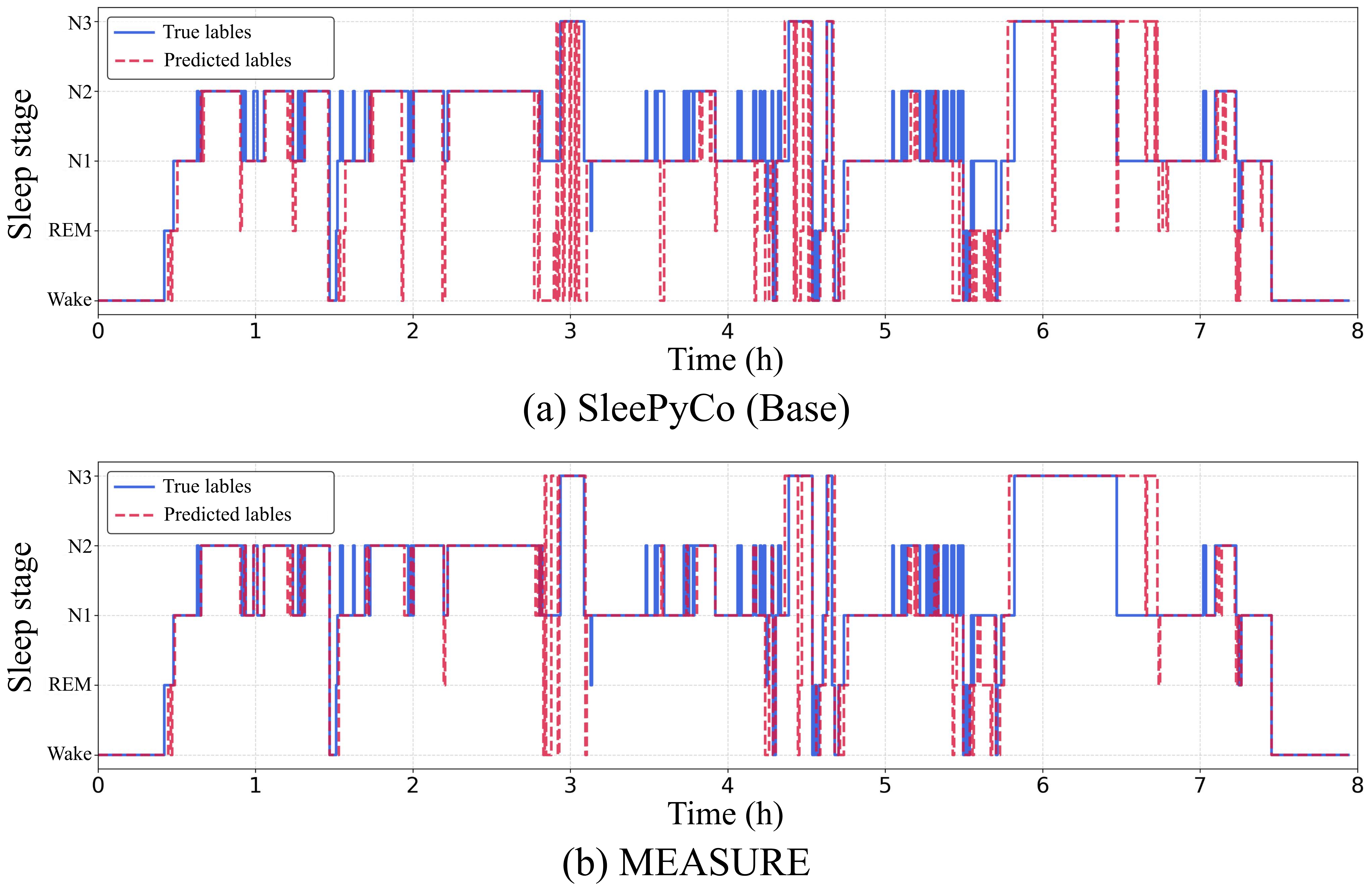} \\
    \caption{Comparison of hypnograms generated by the baseline model (a) SleePyCo and (b) the proposed MEASURE model on SleepEDF-20. True sleep stages (blue) and predicted stages (red dashed) are visualized over time.}
    \label{fig:hypo}
\end{figure}

We conducted a comprehensive evaluation in comparison to SOTA methods for sleep staging, as well as various domain generalization techniques, including IRM (minimizing risk across different environments) \cite{arjovsky2019invariant}, PCL (a proxy-based contrastive learning approach) \cite{yao2022pcl}, and SleepDG (distribution matching of both global and local sleep sequences) \cite{wang2024generalizable}. All DG approaches, except for SleepDG, were trained using the SleePyCo backbone. The comparison was carried out using multiple metrics \cite{sokolova2009systematic}, including accuracy (ACC), macro-averaged F1 score (F1), and Cohen’s Kappa ($\kappa$). Cohen’s Kappa, which adjusts for chance agreement in label predictions, is a crucial metric given the severe class imbalance in sleep staging. As shown in Table \ref{table:comparison} and \ref{table:DG}, our method demonstrated superior performance across both benchmark datasets, SleepEDF-20 and MASS. Table \ref{table:comparison} reports the classification performance aggregated across all folds, following standard sleep staging evaluation protocols. Conversely, Table \ref{table:DG} presents per-fold metrics, adhering to the established evaluation methodology in DG research. 
For the SleepEDF-20 dataset, our approach achieved an accuracy of 87.3$\%$, an F1 score of 81.5$\%$, and a $\kappa$ of 0.826, while for the MASS dataset, it yielded competitive results with an accuracy of 88.3$\%$, an F1 score of 83.6$\%$, and a $\kappa$ of 0.826 in table \ref{table:comparison}.
Table \ref{table:DG} demonstrates that our method achieves the best performance with the lowest standard deviation than other DG methods. 
Fig. \ref{fig:hypo} compares hypnograms from the baseline and MEASURE models, showing that MEASURE aligns more closely with the ground truth, especially in non-wake sleep stages. Experimental results demonstrate the superiority of our method in sleep staging and over other DG approaches.

\begin{table}[t]
    \centering
    \caption{Ablation study on the effects of minimal sufficient learning and multi-scale learning on SleepEDF-20 and MASS datasets.}
    \label{tab:ablation}
    \setlength{\tabcolsep}{4pt}
    \begin{tabular}{ccccc|ccc}
        \toprule
        \multirow{2}{*}{Minimal} & \multirow{2}{*}{Multi} & \multicolumn{3}{c}{SleepEDF-20} & \multicolumn{3}{c}{MASS} \\
        \cmidrule(lr){3-5} \cmidrule(lr){6-8}
         & & $\kappa$ & ACC (\%) & F1 (\%) & $\kappa$ & ACC (\%) & F1 (\%) \\
        \midrule
        \midrule
        \checkmark &  & 0.806 & 85.9 & 79.0 & 0.821 & 87.9 & 83.1 \\
        & \checkmark & 0.808 & 85.9 & 80.1 & 0.821 & 88.0 & 82.8 \\
        \checkmark & \checkmark & \textbf{0.826} & \textbf{87.3} & \textbf{81.5} & \textbf{0.826} & \textbf{88.3} & \textbf{83.6} \\
        \bottomrule
    \end{tabular}
\end{table}

\subsection{Ablation studies}

\textbf{Effect of multi-scale and minimal sufficient representation learning on model performance.} To validate the effectiveness of MEASURE, we conducted ablation studies to assess the individual contributions of multi-scale feature learning and minimal sufficient representation learning. In the absence of minimal sufficient learning, we applied supervised contrastive learning (SCL) in a multi-scale manner as an alternative. Conversely, when multi-scale learning was not applied, our objective was only applied at the features extracted from the last layer. The results are presented in Table \ref{tab:ablation}. The results indicate that neither minimal sufficient learning nor multi-scale learning alone led to significant performance improvements. This suggests that minimal sufficient learning alone may diminish the informativeness of lower-level features, while multi-scale learning alone may be inadequate in preventing the accumulation of domain-relevant information from earlier layers. These findings underscore the necessity of carefully integrating multi-scale learning with minimal sufficient representation learning to leverage their complementary strengths.

\begin{figure}[h] 
    \centering 
    \includegraphics[width=0.48\textwidth]{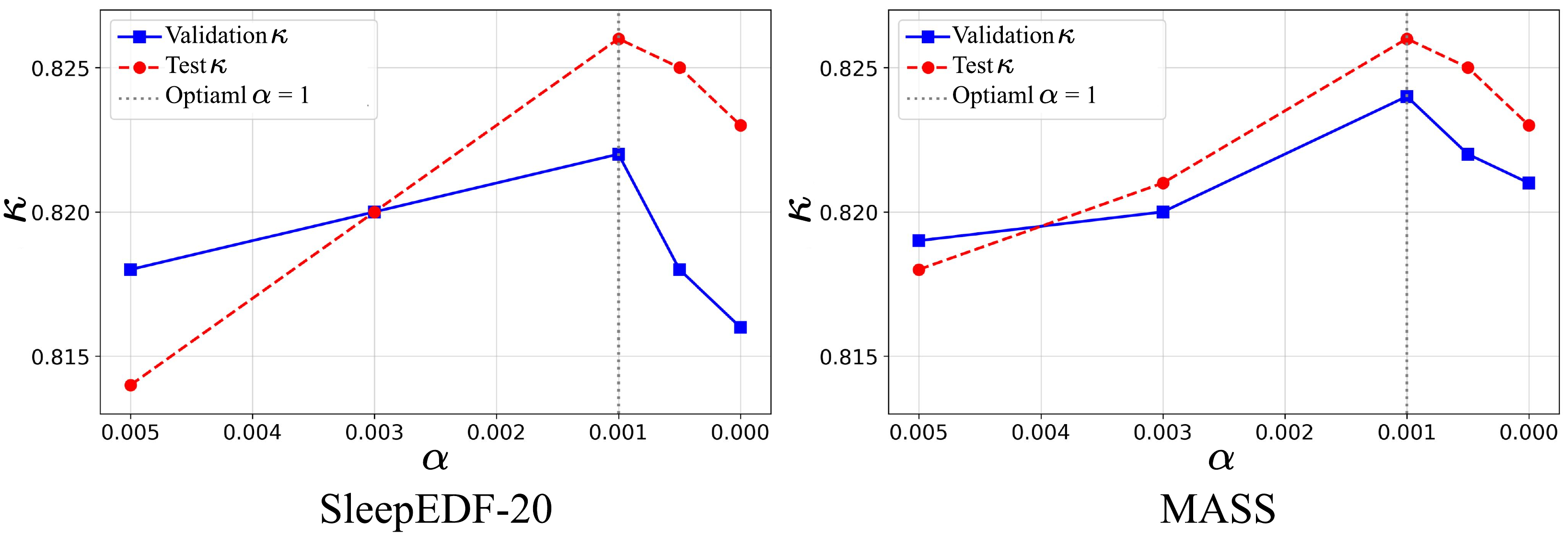} \\
    \caption{Performance comparison across varying the $\alpha$ on SleepEDF-20 and MASS datasets.}
    \label{fig:hyperparmeter}
\end{figure}

\textbf{Analysis of regularization parameter $\alpha$.}
We conducted ablation studies to evaluate the influence of the regularization parameter $\alpha$ on model performance, as illustrated in Fig. \ref{fig:hyperparmeter}. The optimal performance was achieved at $\alpha = 0.001$, indicating that appropriate regularization plays a crucial role in enhancing domain generalization. In contrast, larger values of $\alpha$ led to an overemphasis on $H(\boldsymbol{z}_i| d_i)$, resulting in a failure to capture meaningful features and a subsequent decline in performance.  These results highlight the significance of carefully balancing regularization to ensure the model retains class-relevant information while mitigating the influence of domain biases. 

\begin{table*}[t!]
\centering 
\caption{Performance comparison of feature alignment at different levels on SleepEDF-20 and MASS datasets. We utilized the features extracted from the last encoder layer (high-level), the fourth layer (middle-level), and the third layer (low-level). } 
\label{table: multi-align} 
\begin{tabular}{c ccc ccc ccccc}
\toprule
\multirow{2}{*}{Dataset} & \multirow{2}{*}{High} & \multirow{2}{*}{Middle} & \multirow{2}{*}{Low} & \multicolumn{3}{c}{Overall Metric} & \multicolumn{5}{c}{Per-class F1 ($\%$)} \\
\\
 & & & & $\kappa$ & Acc ($\%$) & F1 ($\%$) & Wake & N1 & N2 & N3 & REM \\ 
\midrule
\midrule

\multirow{4}{*}{SleepEDF-20} 
 & $\checkmark$ & & & 0.806 & 85.8 & 79.0 & \textbf{93.2} & 43.8 & 88.3 & 88.0 & 81.9 \\
 & $\checkmark$ & & $\checkmark$ & 0.811 & 86.2 & 80.5 & 90.5 & 50.1 & 88.3 & 87.9 & 85.8 \\
 & $\checkmark$ & $\checkmark$ & & \textbf{0.825} & \textbf{87.3} & \textbf{81.5} & 92.6 & 50.4 & \textbf{89.3} & \textbf{88.8} & \textbf{86.3} \\
 & $\checkmark$ & $\checkmark$ & $\checkmark$ & 0.816 & 86.5 & 81.1 & 91.5 & \textbf{51.7} & 88.8 & 88.0 & 85.5 \\ 
 
 \midrule

\multirow{4}{*}{MASS} 
 & $\checkmark$ & & & 0.821 & 87.9 & 83.1 & 87.7 & 60.3 & 92.4 & 88.1 & 87.0 \\
 & $\checkmark$ & & $\checkmark$ & 0.817 & 87.7 & 82.4 & 86.7 & 57.6 & 92.4 & \textbf{88.4} & 86.8 \\
 & $\checkmark$ & $\checkmark$ &  & 0.823 & 88.1 & 83.4 & \textbf{88.4} & 60.7 & 92.5 & 88.3 & 87.1 \\
 & $\checkmark$ & $\checkmark$ & $\checkmark$ & \textbf{0.826} & \textbf{88.3} & \textbf{83.6} & 88.2 & \textbf{61.3} & \textbf{92.6} & 88.2 & \textbf{87.6}  \\
\bottomrule
\end{tabular}
\end{table*}

\subsection{Analysis}

\begin{figure}[h]
\centering
    \includegraphics[width=0.45\textwidth]{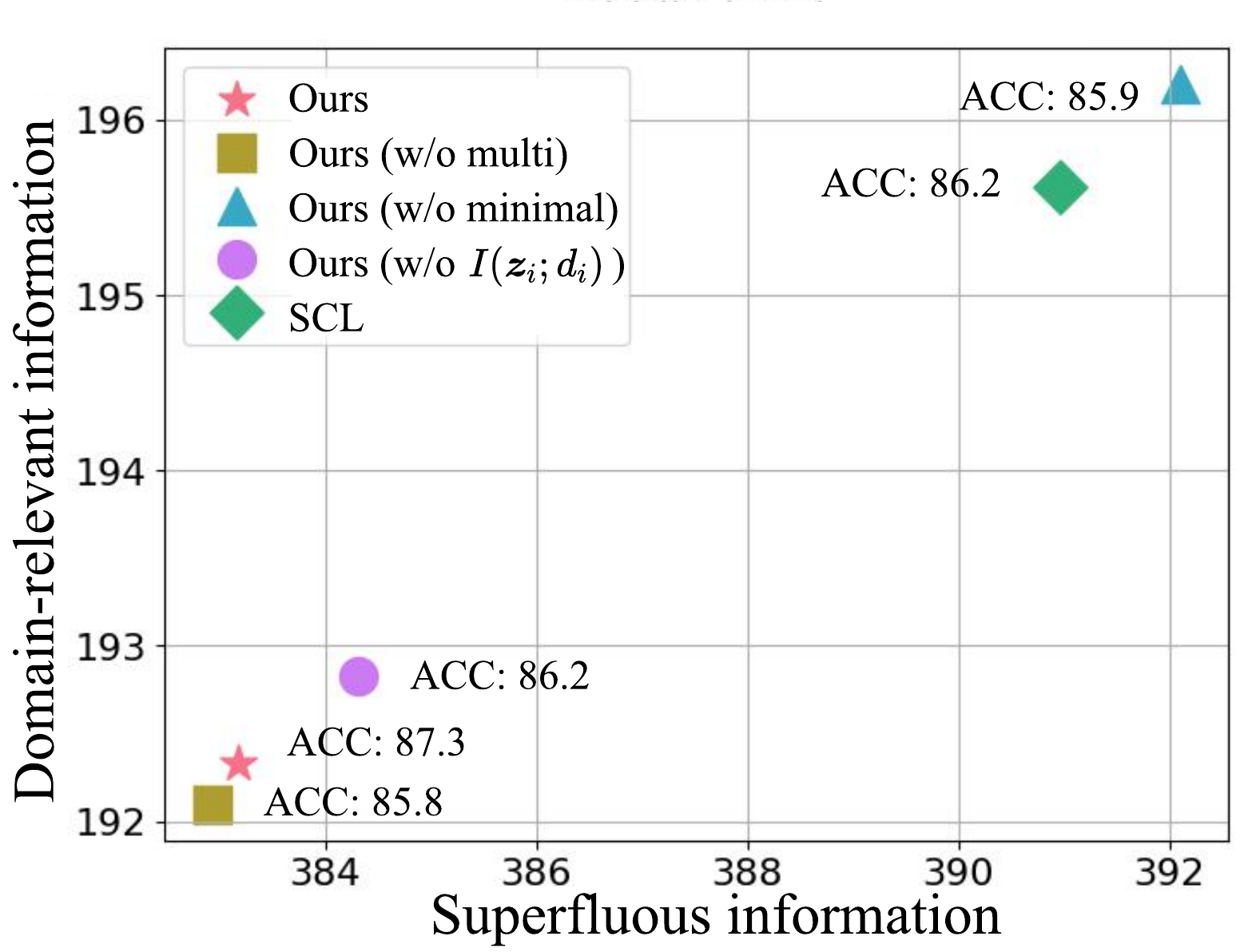} 
    \caption{Visualization of correlation between the superfluous information and domain-relevant information.}
    \label{fig: MI}
\end{figure}

\textbf{Investigation of superfluous and domain-relevant information.} 
To evaluate the effectiveness of our method in reducing superfluous information and capturing domain-invariant features, we conducted an analysis of four different approaches, as illustrated in Fig. \ref{fig: MI}. The information quantities at high-level features depicted in the figure were approximated using the vMF distribution, which is used in our method. Our method achieved the lowest quantities of superfluous information $I(\boldsymbol{z}_i; \boldsymbol{v}_i | \boldsymbol{v}_p)$ and domain-relevant information $I(\boldsymbol{z}_i | d_i)$, effectively minimizing both during training and achieving superior performance. Furthermore, we observed that our method reduced both superfluous and domain-relevant information more effectively than the variant trained without the minimization of $I(\boldsymbol{z}_i | d_i)$ (Ours (w/o $I(\boldsymbol{z}_i | d_i)$)). This result suggests that our method effectively mitigates excess domain-relevant information embedded within superfluous features.

\textbf{Exploring optimal feature alignment levels.} We conducted ablation studies to determine which level of features should be aligned for optimal performance. Among the five encoder layers, we used the output features from the final layer (high-level), the fourth layer (middle-level), and the third layer (low-level). The results of this analysis are presented in Table \ref{table: multi-align}. Our findings reveal that aligning only high-level features leads to a decrease in performance, whereas including other-level feature alignment results in performance improvement. The observed decline is likely due to an overemphasis on the final layer, which prevents proper alignment of features from previous layers. This discrepancy is particularly evident in the model's performance degradation across sleep stages other than the wake stage and is further exacerbated in the SleepEDF-20 dataset, where the wake stage is disproportionately represented. The high-level features are well-suited for capturing high-frequency components, such as the beta rhythm (13–30 Hz), which is characteristic of the wake stage. However, these features exhibit limited diversity, rendering them insufficient for effectively representing other sleep stages. These findings highlight the importance of multi-scale learning, which ensures the preservation and integration of information across different feature hierarchies.

\begin{figure}[h] 
    \centering 
    \includegraphics[width=0.45\textwidth]{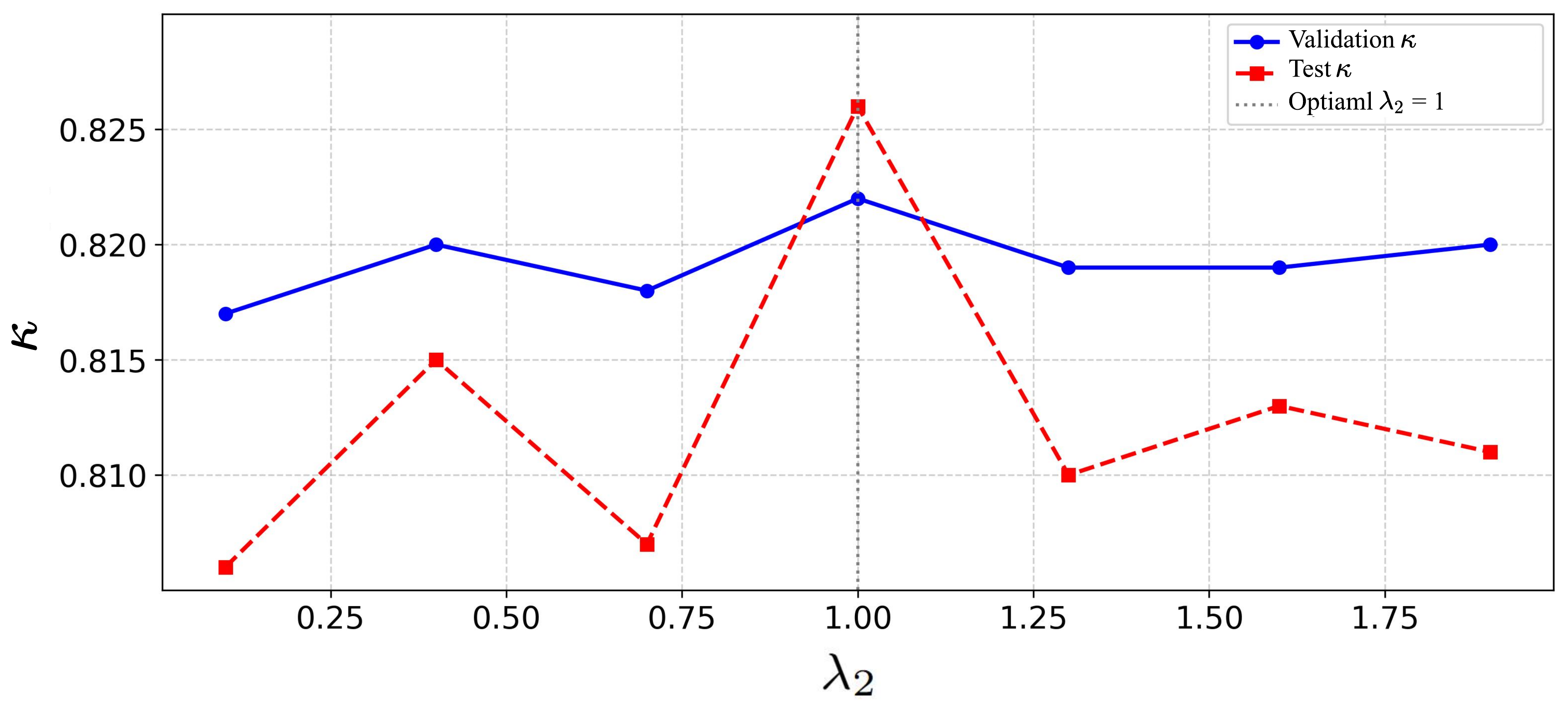} \\
    \caption{Performance results for different values of $\lambda_2$.}
    \label{fig: ab2}
\end{figure}

\textbf{Analysis of $\lambda_2$ in Eq.~(\ref{eq: lagrangian2}).} 
To investigate the influence of $\lambda_2 = 1$ in Eq. (\ref{eq: lagrangian2}), we conducted experiments on the SleepEDF-20 dataset by systematically varying its value. From the experimental results, we observed that setting $\lambda_1 = 1$ yields better performance. The results of these experiments are presented in Fig. \ref{fig: ab2}. In the case where $\lambda_1 > 1$, the coefficient in front of $H(\boldsymbol{z}_i)$ is positive, causing the model to attempt to minimize $H(\boldsymbol{z}_{i})$. Minimizing $H(\boldsymbol{z}_{i})$ reduces the amount of information contained in $\boldsymbol{z}$, which appears to hinder the learning process. Conversely, when $\lambda_1 < 1$, the model simultaneously maximizes both $H(\boldsymbol{z}_{i})$ and $H(\boldsymbol{z}_{i}|d_i)$. While maximizing $H(\boldsymbol{z}_{i}|d_i)$ reduces the domain-relevant information $I(\boldsymbol{z}_{i};d_i)$, maximizing $H(\boldsymbol{z}_{i})$ increases $I(\boldsymbol{z}_{i};d_i)$, as $I(\boldsymbol{z};d) = H(\boldsymbol{z}) -H(\boldsymbol{z}|d)$. Therefore, setting $\lambda_1 = 1$ allows the model to minimize $I(\boldsymbol{z}_{i};d_i)$ by focusing entirely on maximizing $H(\boldsymbol{z}_{i}|d)$, enabling the extraction of more domain-invariant features.

\begin{figure}[h] 
    \centering
    \includegraphics[width=0.5\textwidth]{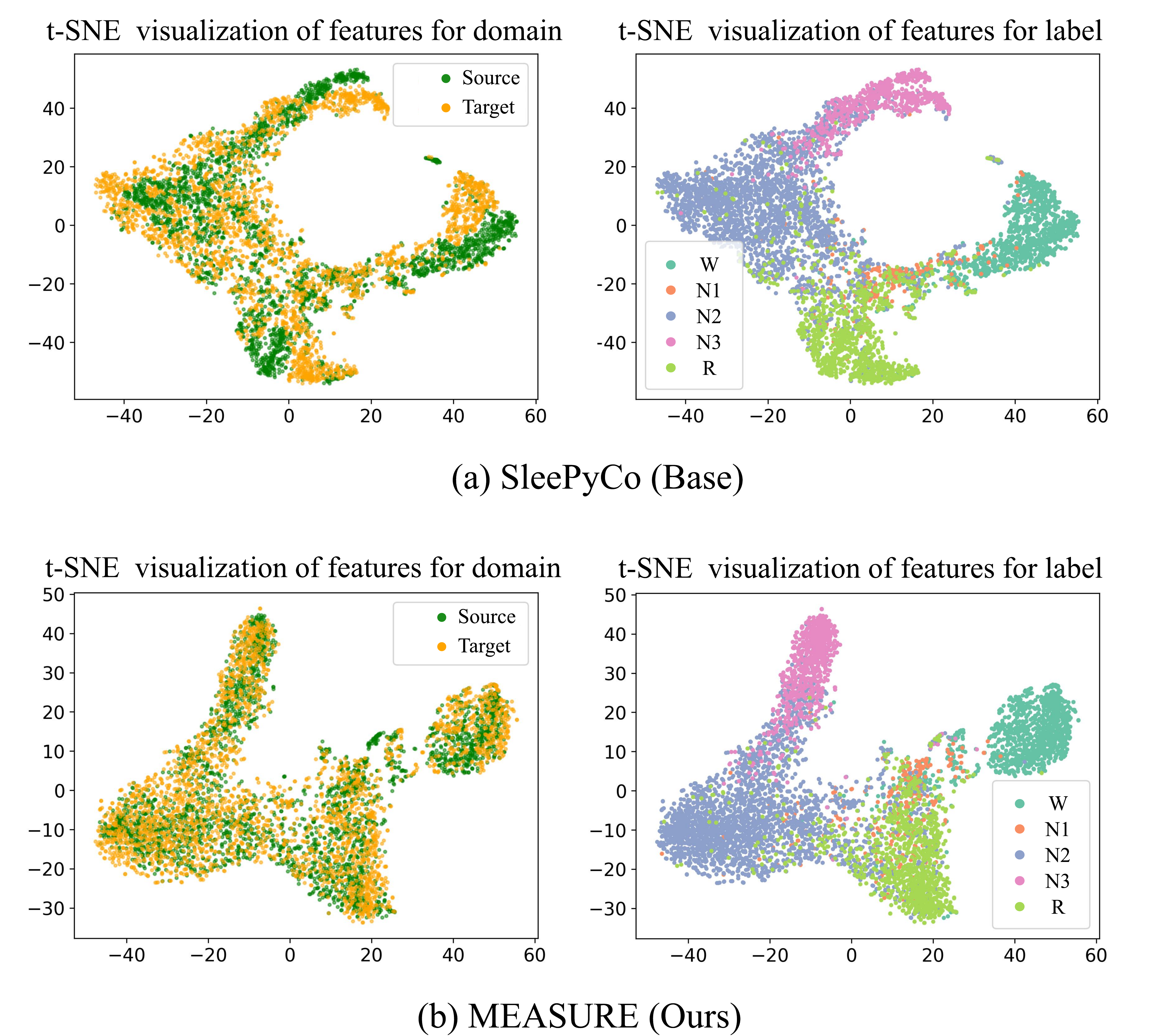}
    \caption{t-SNE visualization of feature distribution on SleepEDF-20, where the source is represented in orange and the target in green.}
    \label{fig:TSNE}
\end{figure}

\textbf{t-SNE visualization.} We performed a feature visualization to further demonstrate the effectiveness of our method, as illustrated in Fig. \ref{fig:TSNE}. The t-SNE visualizations show distributions of features between source (green) and target (orange) domains. For effective visualization in SleepEDF-20, we selected subject 9, which exhibits significant variation, as the target for our analysis. The feature distribution in SleePyCo exhibits misalignment between the source and target domains. In contrast, our MEASURE approach achieves a much more aligned distribution between these domains, indicating that our method effectively generalizes unseen data well.

\section{Discussion and Conclusion}
In this work, we proposed a novel framework, Multi-scalE minimAl SUfficient Representation lEarning (MEASURE), designed to minimize excess domain-relevant information within superfluous features while preserving essential information through the alignment of multi-scale representations. Extensive experiments conducted on publicly available sleep staging datasets demonstrate that our approach consistently outperforms SOTA techniques.

A key theoretical contribution of our work is in demonstrating that reducing excess domain-relevant information ultimately leads to the minimization of the conditional entropy $H(\boldsymbol{z}|d)$. his principle is consistent with traditional methods, such as Domain-Adversarial Neural Networks (DANN)\cite{ganin2016domain}. However, in contrast to existing methods that require additional adversarial training, our framework achieves this without the need for extra modules or training procedures. Furthermore, our ablation studies indicate that while minimal sufficient learning aids in learning domain-invariant features, it leads to overfitting in specific layers. To address this, we propose a novel integration with multi-scale learning, which effectively mitigates these limitations by jointly preserving essential information while maintaining domain invariance. This theoretical insight provides a deeper understanding of the underlying mechanisms, offering valuable guidance for future research endeavors.

\appendix
In this section, we provide the formal proof of the minimal sufficient learning method proposed in the paper. 

\subsection{Proof of Theorem 1} \label{Appendix: a}
\textbf{Theorem} \ref{theorem: 1} The minimal sufficient representation  $\boldsymbol{z}_1^{min}$ is more domain-invariant compared to the sufficient representation $\boldsymbol{z}_1^{suf}$.

\textbf{Proof:}
First, recall that $\boldsymbol{z}_1^{suf}$ is a sufficient representation of $\boldsymbol{v}_1$ with respect to $\boldsymbol{v}_2$, meaning: $I(\boldsymbol{z}_1^{suf} ; \boldsymbol{v}_2) = I(\boldsymbol{v}_1 ; \boldsymbol{v}_2)$. We begin by examining the mutual information between the sufficient representation $\boldsymbol{z}_1^{suf}$ and the domain label $d$:

\begin{align}
I(\boldsymbol{z}_1^{suf}; d) &= H(d) - H(d|\boldsymbol{z}_1^{suf}) \label{eq:I2E}\\
& = H(d) - H(d|\boldsymbol{z}_1^{suf}, \boldsymbol{v}_2) - I(d; \boldsymbol{v}_2|\boldsymbol{z}_1^{suf}) \label{eq:pm_enropies}\\
&= H(d) - H(d|\boldsymbol{v}_2) + H(d|\boldsymbol{v}_2) \nonumber \\
&\quad - H(d|\boldsymbol{z}_1^{suf}, \boldsymbol{v}_2) - I(d; \boldsymbol{v}_2|\boldsymbol{z}_1^{suf}) \label{eq:rearrange}\\
&= I(d; \boldsymbol{v}_2) + I(\boldsymbol{z}_1^{suf}; d|\boldsymbol{v}_2) - I(d; \boldsymbol{v}_2|\boldsymbol{z}_1^{suf})
\label{eq:H2I}\\
&\geq I(d; \boldsymbol{v}_2) + I(\boldsymbol{z}_1^{suf}; d | \boldsymbol{v}_2) - I(d; \boldsymbol{v}_2 | \boldsymbol{z}_1^{min}) \label{eq:data_processing}\\
&= I(\boldsymbol{z}_1^{suf}; d | \boldsymbol{v}_2) + I(\boldsymbol{z}_1^{min}; d) \label{eq:min_suf_def}\\
&\geq I(\boldsymbol{z}_1^{min}; d).
\label{eq:final_inequality}
\end{align}

Here is the explanation for some steps:
\begin{itemize}
    \item Eq.~(\ref{eq:pm_enropies}) We can further decompose this using the chain rule of entropy:
    $I(\boldsymbol{z}_1^{suf}; d) = H(d) - H(d|\boldsymbol{z}_1^{suf}, \boldsymbol{v}_2) - I(d; \boldsymbol{v}_2|\boldsymbol{z}_1^{suf})$. 
    \item Eq.~(\ref{eq:H2I}) Recognizing mutual information terms, we get Eq~(\ref{eq:H2I}):
$I(\boldsymbol{z}_1^{suf}; d) = I(d; \boldsymbol{v}_2) + I(\boldsymbol{z}_1^{suf}; d|\boldsymbol{v}_2) - I(d; \boldsymbol{v}_2|\boldsymbol{z}_1^{suf})$.
    \item Eq.~(\ref{eq:data_processing}) Inequality Eq.~(\ref{eq:data_processing}) is due to the data processing inequality. Since $\boldsymbol{z}_1^{min}$ is a function of $\boldsymbol{z}_1^{suf}$, we have $I(d; \boldsymbol{v}_2 | \boldsymbol{z}_1^{suf}) \leq I(d; \boldsymbol{v}_2 | \boldsymbol{z}_1^{min})$.
    \item Eq.~(\ref{eq:min_suf_def}) For $\boldsymbol{z}_1^{min}$, we have $I(\boldsymbol{z}_1^{min}; d) = I(d; \boldsymbol{v}_2) - I(d; \boldsymbol{v}_2 | \boldsymbol{z}_1^{min})$.
    \item Eq (\ref{eq:final_inequality}) Inequality Eq (\ref{eq:final_inequality}) holds because mutual information is non-negative, so $I(\boldsymbol{z}_1^{suf}; d | \boldsymbol{v}_2) \geq 0$.
\end{itemize}

\subsection{Proof of Eq.~(\ref{eq: lagrangian2})} \label{Appendix:b.1}

The superfluous information $I(\boldsymbol{z}_{i};\boldsymbol{v}_{i}|\boldsymbol{v}_p)$ can be decomposed as:
\begin{align}
I\left(\boldsymbol{z}_i;\boldsymbol{v}_i|\boldsymbol{v}_p\right)&=H(\boldsymbol{z}_i|\boldsymbol{v}_p)-H(\boldsymbol{z}_i|\boldsymbol{v}_i,\boldsymbol{v}_p)\\
&= H(\boldsymbol{z}_i|\boldsymbol{v}_p),
\end{align}
where the conditional entropy $H(\boldsymbol{z}_i|\boldsymbol{v}_i, \boldsymbol{v}_p) = 0$ because $\boldsymbol{z}_i$ is determined given $\boldsymbol{v}_i$ (we used deterministic encoder). Based on the above derivations and Eq. (\ref{eq: 2}), we finally obtain the general objective below:
\begin{align}
    \mathcal{L}(\phi) &=   \lambda_1I(\boldsymbol{z}_i;\boldsymbol{v}_i|\boldsymbol{v}_p) + \lambda_2I(\boldsymbol{z}_i;d_i) -I(\boldsymbol{z}_i;\boldsymbol{v}_{p})
    \\ &= \lambda_1( H(\boldsymbol{z}_{i}|\boldsymbol{v}_{p})) + \lambda_2 (H(\boldsymbol{z}_i)-H(\boldsymbol{z}_i|d_i)) \nonumber \\
    & \quad-H(\boldsymbol{z}_{i}) +H(\boldsymbol{z}_{i}|\boldsymbol{v}_{p}) \\ \label{eq: 25}
    &= (\lambda_1+1)( H(\boldsymbol{z}_{i}|\boldsymbol{v}_{p})) + (\lambda_2 -1)H(\boldsymbol{z}_{i}) -\lambda_2 H(\boldsymbol{z}_i|d_i).
\end{align}

\subsection{Proof of Eq. (\ref{eq: 8})} \label{Appendix:b.2}

The von Mises–Fisher distribution is a widely used probability distribution on the hypersphere. It is expressed as:
\begin{equation}
    p(\boldsymbol{x};\boldsymbol{\mu},\kappa) = C_n(\kappa) \operatorname{exp}(\kappa \boldsymbol{\mu}^T \boldsymbol{x}),
\end{equation}
\begin{equation}
     C_n(\kappa) =  \frac{\kappa^{n/2-1}}{(2\pi)^{n/2}I_{n/{2-1}}(\kappa)},
\end{equation}
where $\boldsymbol{\mu}$ is the mean direction, $\kappa$ denotes the concentration parameter of the vMF distribution, and  $I_n$ denotes the modified Bessel function of the first kind at order $n$. 

The representation $\boldsymbol{z}$ is $\ell_2$-normalized in the hypersphere space. Hence, The variational distribution $q_\phi(\boldsymbol{z}_i|\boldsymbol{v}_p)$ can be adequately approximated by the vMF distribution as, similar to \cite{wen2024mveb}:
\begin{equation}
   q_\phi(\boldsymbol{z}_i|\boldsymbol{v}_p) = C_n(\kappa) \operatorname{exp}(\kappa \boldsymbol{z}_p \cdot \boldsymbol{z}_i). \\
\end{equation}
We assume that $\kappa$ is constant and use $\boldsymbol{z}_p$ as $\boldsymbol{\mu}$.
Hence, Eq. (\ref{eq: 6}) can be reformulated as follows:
\begin{equation}
    H(\boldsymbol{z}_i|\boldsymbol{v}_p) \leq - \mathbb{E}_{p(\mathbf{z}_i,\mathbf{v}_p)}[\kappa \boldsymbol{z}_p^T \boldsymbol{z}_i] -\log C_n(\kappa).
\end{equation}
Eq. (\ref{eq: 7}) can be expressed as follows:
\begin{equation} \label{eq: 23}
    \bar{\mathcal{L}}(\phi) = -\mathbb{E}_{p(\mathbf{z}_i,\mathbf{v}_p)}[ \boldsymbol{z}_p^T \boldsymbol{z}_i] - \beta H(\boldsymbol{z}_i|d_i),
\end{equation}
where $\beta = \frac{1}{(\lambda +1)\kappa}$ is the balance factor.

\subsection{Proof of Eq.~(\ref{eq: cl})} \label{Appendix:b.3}
$\mathbb{E}_{p(\boldsymbol{z}_i,\boldsymbol{z}_p)}[\boldsymbol{z}_i^T \boldsymbol{z}_p]$ can be decomposed using Monte Carlo approximation and empirical distribution as:

\begin{align}
    \mathbb{E}_{p(\boldsymbol{z}_i, \boldsymbol{z}_p)}[\boldsymbol{z}_i^T \boldsymbol{z}_p] &= \sum_{i \in I} \sum_{p\in P(i)} p(\boldsymbol{z}_p|\boldsymbol{z}_i) p(\boldsymbol{z}_i) \: \boldsymbol{z}_i^T \boldsymbol{z}_p \\
    & \approx \frac{1}{|I|} \sum_{i \in I} \sum_{p\in P(i)} \frac{1}{|P(i)|} \: \boldsymbol{z}_i^T \boldsymbol{z}_p, \\
    \mathbb{E}_{p(\boldsymbol{z}_i,\boldsymbol{z}_p)}[\boldsymbol{z}_i^T \boldsymbol{z}_p /\tau] &= \frac{1}{|I|} \sum_{i \in I} \sum_{p\in P(i)} \frac{1}{|P(i)|} \: \boldsymbol{z}_i^T \boldsymbol{z}_p /\tau,
\end{align}
where $I$ refers to the set of indices corresponding to the batch samples.
Eq.~(\ref{eq: 8}) can rewrite as follows: 
\begin{equation} \label{eq:aa}
        \hat{\mathcal{L}}(\phi) / \tau = - \frac{1}{|I|} \sum_{i \in I} \sum_{p\in P(i)} \frac{1}{|P(i)|} \: \boldsymbol{z}_i^T \boldsymbol{z}_p /\tau -\beta /\tau H(\boldsymbol{z}_i|d_i).
\end{equation}
We can rewrite Eq. (\ref{eq:aa}) as follows:
\begin{align}
        \hat{\mathcal{L}}_{\mathrm{w/ neg}}(\phi) / \tau =& -\frac{1}{|I|} \sum_{i \in I} \frac{1}{|P(i)|} \sum_{p \in P(i)} \log \exp(\frac{\boldsymbol{z}_{i}^T \boldsymbol{z}_{p}}{{\tau}})   \nonumber \\
        &-\beta /\tau H(\boldsymbol{z}_i|d_i).\label{eq: 39}
\end{align}
We also consider a set of negative pairs as follows:
\begin{align}
    \tilde{\mathcal{L}}(\phi) &= -\sum_{i \in I} \frac{1}{|P(i)|} \sum_{p \in P(i)} \log \frac{\exp(\frac{\boldsymbol{z}_i^T \boldsymbol{z}_p} {\tau})}{ 
\sum_{n \in N(i)} \exp (\frac{\boldsymbol{z}_i^T \boldsymbol{z}_n} {\tau})} \nonumber \\
& \quad-\alpha  H(\boldsymbol{z}_i|d_i),
\end{align}
where $\alpha$ is the regularization parameter.

\subsection{Computation of Entropy } \label{Appendix:entropy}
We follow the derivation from \cite{wen2024mveb}, with the key difference being that it is conditioned on the given domain label $d$. The gradient of $H(\boldsymbol{z}|d)$ w.r.t. $\phi$ can be decomposed as:
\begin{align}
\nabla_{\boldsymbol{\phi}} H(\boldsymbol{z}|d) = -\nabla_{\boldsymbol{\phi}} \mathbb{E}_{q_{\boldsymbol{\phi}}(\boldsymbol{z},d)}[\log q(\boldsymbol{z}|d)] \nonumber
\\ \quad -\mathbb{E}_{q(\boldsymbol{z},d)}[\nabla_{\phi} \log q_{\boldsymbol{\phi}}(\boldsymbol{z}|d)],
\end{align}
where $q(\boldsymbol{z},d)$ without the subscript $\phi$ means the gradient of computation is irrelevant to $\phi$. The second term can be further decomposed as:
\begin{align}\mathbb{E}_{q(\boldsymbol{z},d)}[\nabla_\phi\log q_\phi(\boldsymbol{z}|d)]&=\mathbb{E}_{q(\boldsymbol{z})}\left[\nabla_\phi q_\phi(\boldsymbol{z}|d)\times\frac1{q(\boldsymbol{z}|d)}\right]\\
&=\nabla_\phi\int q_\phi(\boldsymbol{z}|d)d\boldsymbol{z}=0.
\end{align}
Hence, we have
\begin{equation}
    \nabla_\phi H(\boldsymbol{z}|d)=-\nabla_\phi\mathbb{E}_{q_\phi(\boldsymbol{z},d)}[\log q(\boldsymbol{z}|d)].
\end{equation}

We adopt the reparameterization trick to address non-differentiable $H(\boldsymbol{z}|d_i)$ w.r.t $\phi$. We introduce the deterministic function $f_\phi$ and any joint distribution $p(\cdot)$ that is independent to model parameter $\phi$.
\begin{equation}
    \boldsymbol{z} = f_{\phi}(\boldsymbol{v}|d)\quad\mathrm{~with~}\quad\boldsymbol{v}\sim p(\boldsymbol{v},d).
\end{equation}
The conditional entropy gradient estimator is eventually derived as follows: \begin{align}
    \nabla_\phi H(\boldsymbol{z}|d) &=-\nabla_\phi\mathbb{E}_{q_\phi(\boldsymbol{z},d)}[\log q(\boldsymbol{z}|d)] \\
    &=-\mathbb{E}_{p(\boldsymbol{v},d)}[\nabla_\phi\log q(f_\phi(\boldsymbol{v}|d))] \\
    &=-\mathbb{E}_{p(\boldsymbol{v},d)}[\nabla_{\boldsymbol{z}}\log q(\boldsymbol{z}|d)\nabla_\phi f_\phi(\boldsymbol{v}|d)],
\end{align}
where $\nabla_{\boldsymbol{z}}\log q(\boldsymbol{z}|d)$ is the score function. $\nabla_\phi f_\phi(\boldsymbol{v}|d)$ can be obtained by direct back-propagation.
We use Stein gradient estimation \cite{li2017gradient} to approximate the score function $\nabla_{\boldsymbol{z}}\log q(\boldsymbol{z}|d)$ as $\hat{\mathbf{G}}^{\mathrm{Stein}}$.
Based on this approximation, the entropy gradient estimator is formulated as:
\begin{align} 
 \nabla_\phi H(\boldsymbol{z}|d) &= -\sum_{d = 1}^{M}\mathbb{E}_{p(\boldsymbol{v}|d)}[\nabla_{\boldsymbol{z}}\log q(\boldsymbol{z}|d)
 \nabla_\phi f_\phi(\boldsymbol{v}|d)] \\ &\approx -\sum_{d = 1}^{M} \mathbb{E}_{p(\boldsymbol{v}|d)}[\hat{\mathbf{G}}^{\mathrm{Stein}}_m \nabla_\phi f_\phi(\boldsymbol{v}|d)] \label{eq:approx_entropy}
 \end{align}
where, $\hat{\mathbf{G}}^{\mathrm{Stein}}_m$ represent the approximation of the score function $\nabla_{\boldsymbol{z}}\log q(\boldsymbol{z}|d)$ computed for the $m$-th domain. $H(\boldsymbol{z}|d)$ can be alternatively represented as
$-\sum_{d= 1}^{M} \mathbb{E}_{p(\boldsymbol{v}|d)}[\hat{\mathbf{G}}^{\mathrm{Stein}}_m \boldsymbol{z}]$ in decent gradient optimization. This is because its gradient, $-\sum_{d = 1}^{M} \mathbb{E}_{p(\boldsymbol{v}|d)}[\hat{\mathbf{G}}^{\mathrm{Stein}}_m \nabla_\phi f_\phi(\boldsymbol{v}|d)]$, provides an approximation of $\nabla_\phi H(\boldsymbol{z}|d)$, as described in Eq.~(\ref{eq:approx_entropy}).

\bibliographystyle{IEEEtran}
\bibliography{ref}

\vfill

\end{document}